\documentclass{article}

\usepackage{microtype}
\usepackage{graphicx}
\usepackage{subfigure}
\usepackage{booktabs} 

\usepackage{hyperref}

\usepackage[accepted]{icml2024}

\usepackage{amsmath}
\usepackage{amssymb}
\usepackage{mathtools}
\usepackage{amsthm}

\usepackage[capitalize,noabbrev]{cleveref}

\usepackage{graphicx}
\usepackage{caption}
\usepackage{pifont}
\usepackage{rotating}
\usepackage{float}

\usepackage{colortbl}  
\usepackage{array}   

\usepackage{xcolor}

\usepackage{multirow}
\usepackage{bbm, dsfont}
\usepackage{colortbl}

\usepackage{titletoc}
\usepackage{etoc}

\theoremstyle{plain}

\theoremstyle{definition}

\theoremstyle{remark}

\usepackage[textsize=tiny]{todonotes}

\icmltitlerunning{Connecting the Dots: Collaborative Fine-tuning for Black-Box Vision-Language Models}

\begin{document}

\twocolumn[
\icmltitle{Connecting the Dots: Collaborative Fine-tuning for \\Black-Box Vision-Language Models}




\begin{icmlauthorlist}
\icmlauthor{Zhengbo Wang}{ustc,casia}
\icmlauthor{Jian Liang}{casia,ucas}
\icmlauthor{Ran He}{casia,ucas}
\icmlauthor{Zilei Wang}{ustc}
\icmlauthor{Tieniu Tan}{casia,ucas,nju}
\end{icmlauthorlist}

\icmlaffiliation{ustc}{University of Science and Technology of China}
\icmlaffiliation{casia}{NLPR \& MAIS, Institute of Automation, Chinese Academy of Sciences}
\icmlaffiliation{ucas}{School of Artificial Intelligence, University of Chinese Academy of Sciences}
\icmlaffiliation{nju}{Nanjing University}

\icmlcorrespondingauthor{Jian Liang}{liangjian92@gmail.com}

\icmlkeywords{Machine Learning, ICML}

\vskip 0.3in
]



\printAffiliationsAndNotice{}  

\begin{abstract}
With the emergence of pretrained vision-language models (VLMs), considerable efforts have been devoted to fine-tuning them for downstream tasks.
Despite the progress made in designing efficient fine-tuning methods, such methods require access to the model's parameters, which can be challenging as model owners often opt to provide their models as a black box to safeguard model ownership.
This paper proposes a \textbf{C}ollabo\textbf{ra}tive \textbf{F}ine-\textbf{T}uning (\textbf{CraFT})
approach for fine-tuning black-box VLMs to downstream tasks, where one only has access to the input prompts and the output predictions of the model.
CraFT comprises two modules, a prompt generation module for learning text prompts and a prediction refinement module for enhancing output predictions in residual style.
Additionally, we introduce an auxiliary prediction-consistent loss to promote consistent optimization across these modules.
These modules are optimized by a novel collaborative training algorithm.
Extensive experiments on few-shot classification over 15 datasets demonstrate the superiority of CraFT. 
The results show that CraFT achieves a decent gain of about 12\% with 16-shot datasets and only 8,000 queries.
Moreover, CraFT trains faster and uses only about 1/80 of the memory footprint for deployment, while sacrificing only 1.62\% compared to the white-box method. 
Our code is publicly available at \url{https://github.com/mrflogs/CraFT}.
\end{abstract}

\section{Introduction}
\label{sec:intro}

In recent years, large-scale pretrained vision-language models have garnered much attention. 
By establishing a link between images and natural language, these models
exhibit impressive zero-shot capabilities and remarkable transfer ability~\cite{radford2021learning, jia2021scaling, alayrac2022flamingo, li2022blip}, demonstrating potential in learning open-world concepts.
One of the most successful large-scale pretrained vision-language models is CLIP~\cite{radford2021learning}. 
After pretraining, CLIP~\cite{radford2021learning} can perform zero-shot recognition by merely providing the class names.
The classification weights are generated by the language encoder through prompting~\cite{liu2023pre}. 

Besides its remarkable zero-shot ability, recent studies have found that CLIP~\cite{radford2021learning} also possesses astonishing transfer ability~\cite{zhou2022learning, zhang2022tip, lu2022prompt}. 
For example, CoOp~\cite{zhou2022learning} can achieve a 15\% improvement compared to zero-shot CLIP~\cite{radford2021learning} with only 16 samples per class by fine-tuning a mere 16k parameters.
However, these methods assume we have access to the model parameters, which is unrealistic in the current era.
Training large vision-language models typically requires extensive computational resources and data, thus leading to high training costs.
Therefore, model owners seldom release the model and the weights to protect the model ownership.
Typically, model owners deploy the models as a service, such as GPT-4~\cite{openai2023gpt4}, where we can only obtain the input and output.
Thus, it is crucial to explore ways to fine-tune powerful vision-language models in the black-box scenario.

To address the aforementioned challenge, we propose \textbf{C}ollabo\textbf{ra}tive \textbf{F}ine-\textbf{T}uning (\textbf{CraFT}), a parameter- and data-efficient fine-tuning approach for black-box VLMs.
The CraFT framework comprises three key components. 
Firstly, it incorporates a prompt generation module designed to learn global text prompts tailored to downstream datasets. 
Given the unavailability of gradients from the black-box model, we leverage derivative-free optimization (DFO) for the module, inspired by prior works~\cite{sun2022black, sun2022bbtv2}. 
The DFO method assumes that the module's parameters adhere to a parameterized distribution.
By sampling solutions from this distribution and calculating the corresponding loss values, it iteratively updates the distribution parameters.
Later, we can obtain the module's parameter from the distribution.
To accelerate the optimization process, the text prompts are projected into a lower-dimensional subspace using a random matrix, as~\citet{aghajanyan2021intrinsic} demonstrates a low-dimensional subspace can be as effective as the full parameter space for fine-tuning.

Secondly, CraFT introduces a prediction refinement module aimed at enhancing the VLM's output predictions. 
This module builds upon the predictions of black-box models which thus can be optimized through gradient descent. 
It consists of a three-layer MLP that learns the prediction's residual, where the residual connection plays a pivotal role in the collaborative training algorithm discussed below.

Thirdly, CraFT develops a novel collaborative training algorithm to optimize the aforementioned modules jointly. 
Given that the prompt generation module and the prediction refinement module are optimized using different optimizers (derivative-free and derivative-based), their joint training poses a challenge.
To address this, we demonstrate that the model with residual connections can be reframed as the addition of outputs of each layer, enabling the modules to be optimized alternately.
Fortunately, both VLMs and the prediction refinement module incorporate shortcut connections, facilitating this iterative optimization. 
To improve training stability, we introduce a prediction-consistent loss that penalizes deviations between the black-box model's output and the refinement module's output.

Our main contributions are summarized as follows:
\begin{itemize}
    \item This paper is among the pioneering works in exploring efficient fine-tuning methods for black-box vision-language models, providing a new framework for fine-tuning black-box VLMs.
    \item CraFT comprises a prompt generation module and a prediction refinement module, which are designed to learn the text prompts and refine the output predictions, respectively. 
    In addition, a collaborative training algorithm and a prediction-consistent loss are proposed to train these modules jointly and collaboratively.
    \item CraFT significantly outperforms black-box baselines on 15 datasets on few-shot classification. 
    Compared to the white-box method, CraFT trains faster and requires only 1/80 of the memory footprint for deployment.
\end{itemize}

\section{Related Work}
\textbf{Vision-Language Models.}
In recent years, vision-language models (VLMs) have gained popularity as fundamental models that aim to connect the modalities of vision and language.
These models are pretrained on large-scale image-text datasets, which endows them with powerful transferable abilities such as zero-shot learning, few-shot adaptation, and in-context learning~\cite{radford2021learning, kim2021vilt, lu2019vilbert, su2019vl, jia2021scaling, wang2023large, chen2023vlp, wen2023editorial}.
Contrastive-based methods have become the mainstream approach in this field.
These methods, including CLIP~\cite{radford2021learning} and ALIGN~\cite{jia2021scaling}, are trained on large-scale web-based noisy image-text pairs. 
They employ a language encoder and a vision encoder to encode the texts and images, respectively, and learn to align their representations through contrastive loss.

\textbf{Efficient Fine-tuning for VLMs.}
Inspired by the prior works in NLP, recent researches focus on developing efficient fine-tuning methods for VLMs on downstream tasks~\cite{zhou2022learning, zhou2022conditional, zhang2022tip, gao2021clip, lu2022prompt, chen2022prompt, derakhshani2022variational, wang2023improving, wang2024baseline}.
Existing efficient fine-tuning methods can be classified into two categories: prompt tuning~\cite{zhou2022learning, zhou2022conditional, lu2022prompt, chen2022prompt} and adapter-style tuning~\cite{gao2021clip, zhang2022tip}.
Prompt tuning methods propose to learn soft text prompts for downstream tasks through back-propagation on few-shot datasets.
For instance, CoOp~\cite{zhou2022learning} proposes to learn soft text prompts through back-propagation on few-shot datasets.
Adapter-style tuning methods, on the other hand, maintain the original zero-shot classifier but refine the output representation.
CLIP-Adapter~\cite{gao2021clip} proposed to add MLPs to refine the visual and text features via a residual connection.
Although these methods have achieved satisfactory results on downstream datasets, they all assume that the entire parameters of VLMs are available. 
However, to safeguard the model ownership, it's difficult to obtain the parameters and architecture of the models.
Therefore, it is necessary to investigate ways to fine-tune black-box VLMs.

\begin{figure*}[t]
    \centering
    \includegraphics[width=0.8\textwidth]{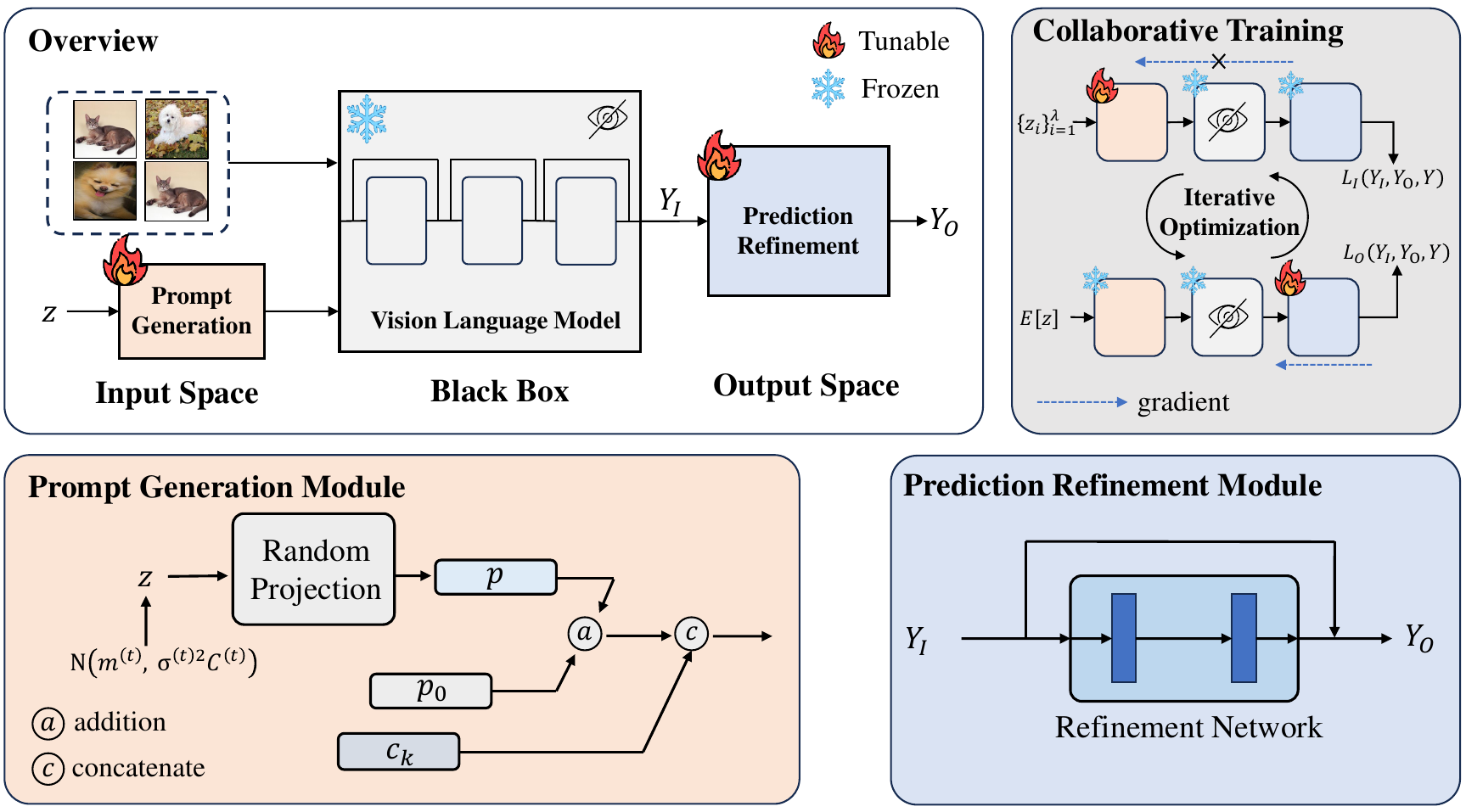}
    \caption{
        \textbf{The architecture of our proposed method.}
        Our proposed method consists of two modules: a prompt generation module and a prediction refinement module.    
        The prompt generation module utilizes the CMA-ES optimizer to learn the text prompts. 
        Specifically, given $z\in\mathbb{R}^{d_0}$ from the CMA-ES optimizer, we project it into the prompt space using a random matrix $A\in\mathbb{R}^{nd\times d_0}$ and add it to initial prompt embeddings $p_0\in\mathbb{R}^{n\times d}$ (e.g., ``a photo of a"). We then concatenate it with the class name embedding $c_k$ to obtain the final prompts $p_k = [p_0 + Az, c_k]$, where $k=1,2,\dots, K$ for $K$ classes.
        The prediction refinement module refines the output of the black-box model using a refinement network in residual style. It can be optimized using gradient descent.
        Since the modules use different optimizers, we propose a novel algorithm to train them collaboratively. 
        For more details on the training process, please refer to section~\ref{subsec:algorithm}. 
   }
    \label{fig:architecture}
\end{figure*}

\textbf{Black-Box Optimization.}
Recently, influenced by developments in NLP, there has been a growing focus on addressing the challenge of fine-tuning VLMs in black-box scenarios~\cite{oh2023blackvip, sun2022bbtv2, sun2022black, yu2023black, malladi2023fine, guo2023black, ouali2023black}.
The methods typically can be split into two categories: zeroth-order optimization and evolutionary algorithms.
Zeroth-order methods~\cite{oh2023blackvip, guo2023black} address the black-box optimization by estimating the gradients, while the evolutionary methods~\cite{sun2022black, yu2023black} solve it by generating candidate solutions through a parameterized distribution.
Though there are some works in black-box VLMs, the underlying assumptions of these methods may be deemed somewhat unreasonable.
For example, LFA~\cite{ouali2023black} assumes that the VLM is dual-towered and we can obtain the corresponding features, which restricts its applicability.
And~\citet{yu2023black} assumes they can access the architecture of VLMs, but update it without gradients. 
In our setting, we assume the model is invisible, and no assumptions are made about its architecture.
Detailed comparison of black-box setting can refer to appendix~\ref{appendix:setting}.

\section{Method}
\label{sec:method}
As depicted in Figure~\ref{fig:architecture}, we divide the model into three distinct parts: the input space, the black-box vision-language model, and the output space.
Since we lack access to the parameters of the black-box model, we can solely integrate learnable modules in the input and output spaces.
In the input space, we propose a prompt generation module, which learns global text prompts for downstream tasks using the Covariance Matrix Adaptation Evolution Strategy (CMA-ES)~\cite{hansen2003reducing}.
In the output space, we propose a prediction refinement module to refine the output prediction of the black-box model in residual style.
Later, we propose a collaborative training algorithm to train them jointly, despite they utilize different optimizers.

\subsection{Prompt Generation Module}
We consider a black-box vision-language model, denoted by $f$.
Given the model, we can only obtain its input and output prediction $f(\{t_k\}_{k=1}^K, \{i_n\}_{n=1}^N)\in\mathbb{R}^{N\times K}$. 
Here, $\{i_n\}_{n=1}^N$ refers to the $N$ images that are uploaded to the black-box vision-language model, and $\{t_k\}_{k=1}^K$ denotes the $K$ class text prompts, where each prompt $t_k$ consists of a predefined text embedding $p_0$ (e.g., ``a photo of a") and a corresponding class name $c_k$. Specifically, we have $t_k = [p_0, c_k]$, $k=1,2,\dots,K$ for $K$ classes.

As shown in Figure~\ref{fig:architecture}, we propose to learn global prompts $p\in\mathbb{R}^{n\times d}$ for the black-box model, where $n$ and $d$ represent the length of the prompts and their dimension, respectively.
Previous work~\cite{aghajanyan2021intrinsic} reveals that a low-dimensional subspace can be as effective as the full parameter space for fine-tuning, we further reduce the search space for fast training by mapping the prompts $p$ into a low-dimensional subspace using a random matrix, i.e., $p = Az$. 
Here, $A\in\mathbb{R}^{nd\times d_0}$ is a random matrix sampled from a Gaussian distribution, and $d_0 \ll nd$ is the dimension of the subspace.
Next, we add the prompts to the initial prompt embeddings $p_0$.
Thus, the optimization can be formulated as follows:
\begin{equation}
    \min_{z} \mathcal{L}(f(\{[p_0 + Az, c_k]\}_{k=1}^K, \{i_n\}_{n=1}^N), Y),
\end{equation}
where $\mathcal{L}$ is the cross-entropy, and $Y$ denotes the ground-truth.
Since the model's gradients are not accessible, we solve this problem using a DFO method, CMA-ES~\cite{hansen2003reducing}.

CMA-ES~\cite{hansen2003reducing} is a parameterized search distribution model that uses a multivariate normal distribution.
At each iteration, CMA-ES generates a population of new query solutions by sampling from the multivariate normal distribution:
\begin{equation}
    z_i \sim m^{(t)} + \sigma^{(t)}\mathcal{N}(0, C^{(t)}), \quad i=1,2,\dots,\lambda.
\end{equation}
Here, $i$ denotes the index of the sampled solution, $\lambda$ is the population size, $m^{(t)}$ represents the distribution mean, $\sigma^{(t)} \ge 0$ is the step-size, and $C^{(t)}$ denotes the covariance matrix of the distribution.
The parameters $m^{(t)}, \sigma^{(t)}, C^{(t)}$ are updated in each iteration to minimize the loss of the sample solutions.
\vspace{-10pt}
\subsection{Prediction Refinement Module}
\vspace{-5pt}
Besides learning text prompts, we further build a refinement network on top of the output of the black-box vision-language model, which learns to refine the output prediction in residual style.\footnote{
The residual connection is necessary for collaborative training algorithm~\ref{subsec:algorithm}.
Table~\ref{tab:refinement} shows its ablation.
}
Specifically, given the original output $Y_I \in \mathbb{R}^{K}$ of the black-box model, the refinement network learns to generate the residual $R(Y_I)$, which is added to the original output to obtain the final result: 
\begin{equation}
    Y_O = Y_I + R(Y_I).
\end{equation}
The refinement network is trained to minimize the cross-entropy.
As the refinement network is built on top of the black-box model and does not require gradients from it, we utilize gradient descent to optimize the refinement network.
\vspace{-10pt}
\subsection{Collaborative Training Algorithm}
\vspace{-5pt}
\label{subsec:algorithm}
As shown in Figure~\ref{fig:architecture}, the prompt generation module and prediction refinement module uses different optimizers (CMA-ES and AdamW, respectively). 
Therefore, optimizing the modules jointly becomes a challenge.
To address this issue, we propose a collaborative training algorithm for them.

Previous works~\cite{mei2016competitive, kandasamy2015high, sun2022bbtv2} have shown that networks with shortcut connections can be decomposed into some additive form.
Therefore, different layers can be optimized separately.
For example, considering the optimization of the first and third layers in a three-layer model with residual connections (similar to our scenario), this optimization problem can be decomposed as follows:
\begin{equation}
\label{eq:decompose}
\begin{aligned}
    \min_{\theta_1, \theta_3} f(x) 
    & = \min_{\theta_1, \theta_3} f_3(x_3) + x_3 \\
    & = \min_{\theta_1, \theta_3} f_3(x_3) + f_2(x_2) + x_2 \\
    & = \min_{\theta_1, \theta_3} f_3(x_3) + f_2(x_2) + f_1(x_1) + x_1 \\
    & = \min_{\theta_3} f_3(x_3) + f_2(x_2) + \min_{\theta_1} f_1(x_1) + x_1,
\end{aligned}
\end{equation}
where $f_i(\cdot)$ denotes the $i$-th layer and its parameter is $\theta_i$, and $x_i$ is the input of the $i$-th layer.
\begin{algorithm}[t]
    \caption{Collaborative Training for CLIP}
    \label{algorithm}
\begin{algorithmic}[1]
    \REQUIRE 
    \begin{minipage}[t]{0.6\textwidth}
        Budget of API calls $\mathcal{B}$, 
        Population size $\lambda$, \\
        Dataset size $|\mathcal{D}|$, 
        Batch size $B$, \\
        Refinement network $R$ with residual connections.
    \end{minipage}
    \STATE Initialize random projections $A$
    \STATE Initialize parameters $m^{(0)}, \sigma^{(0)}, C^{(0)}$ 
    \FOR {$i = 1$ to $\mathcal{B} / \lambda$}
        \STATE \textcolor{blue!50!cyan}{\# Optimize prompt generation module}
        \STATE Sample $\lambda$ solutions $z_i \sim m^{(t)} + \sigma^{(t)}\mathcal{N}(0, C^{(t)})$ 
        \STATE Compute the fitnesses using Equation~(\ref{prompt_equation})
        \STATE Update $m^{(t)}, \sigma^{(t)}, C^{(t)}$ using the CMA-ES
        \STATE \textcolor{blue!50!cyan}{\# Optimize prediction refinement module}
        \FOR{$j=1$ to $|\mathcal{D}| / B$}
            \STATE Sample batch $(Y_I, Y)$
            \STATE Compute the refined output $Y_O = Y_I + R(Y_I)$
            \STATE Compute the loss using Equation~(\ref{refine_equation})
            \STATE Update refinement network $R$ using AdamW
        \ENDFOR
    \ENDFOR
    \STATE \textbf{return} prompts $p = \mathbb{E}_z[p_0 + Az]$ and network $R$
\end{algorithmic}
\end{algorithm}  

\noindent
Thus, based on Eq.~(\ref{eq:decompose}), with residual connections, $f_3$ and $f_1$ can be optimized independently.
Fortunately, vision-language models typically have residual connections, and the prediction refinement network also comprises a shortcut connection. 
Therefore, we can iteratively optimize the prompt generation and the prediction refinement module.

Moreover, to enhance training stability, we further propose a prediction-consistent loss.
Specifically, we use an additional Kullback–Leibler divergence to constrain the output of the black-box model and the refinement module.
Thus, the loss for the CMA-ES optimizer is formulated as follows:
\begin{equation}
    \label{prompt_equation}
    \mathcal{L}_I = CE(Y_I, Y) + \lambda_I * KL(Y_I\|Y_O), 
\end{equation}
where $\lambda_I$ is a hyper-parameter, $Y_I$ denotes the output of the black-box model, $Y_O$ is the output of the refinement network, $Y$ represents the ground-truth label, $CE$ is the cross-entropy loss, and $KL$ is the Kullback-Leibler divergence.
Similarly, during the optimization of the prediction refinement network, we also add a KL divergence loss, which serves as the regularization term, for training stabilization.
The objective for the prediction refinement module can be written as follows:
\begin{equation}
    \label{refine_equation}
    \mathcal{L}_O = CE(Y_O, Y) + \lambda_O * KL(Y_O\|Y_I),
\end{equation}
where $\lambda_O$ is a hyper-parameter.
Thus, the term ``collaboratively" implies the algorithm can jointly optimize two modules while ensuring they work together through consistency loss rather than interfering with each other due to the sequential nature of the modules.

\begin{figure*}[!htbp]
    \centering
    \begin{minipage}{0.245\textwidth}
        \centering
        \includegraphics[width=1.13\textwidth]{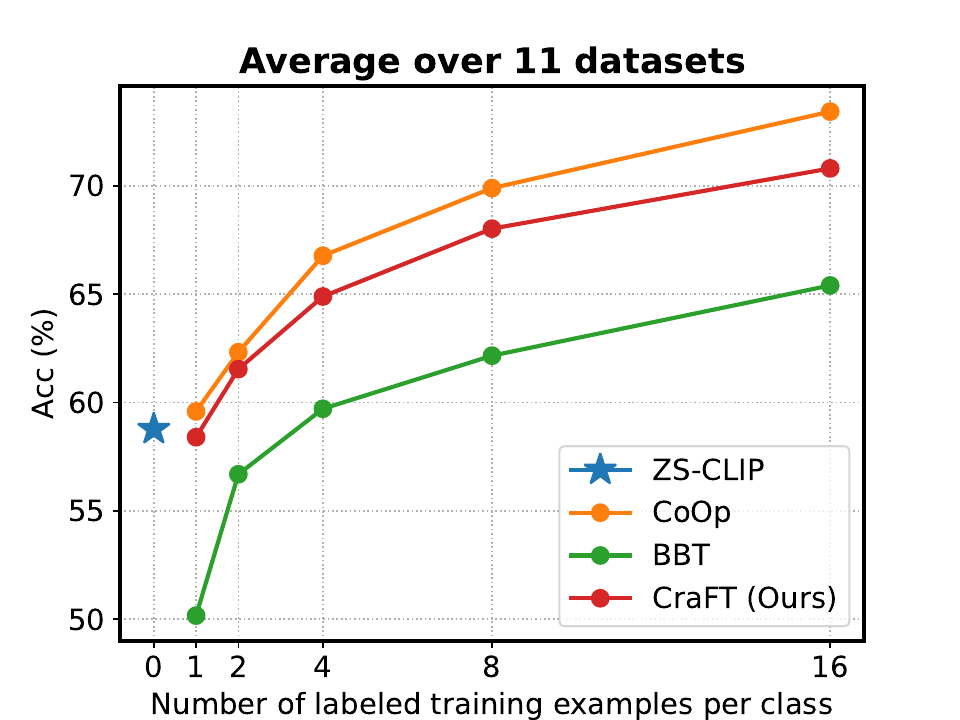}
    \end{minipage}
    \begin{minipage}{0.245\textwidth}
        \centering
        \includegraphics[width=1.13\textwidth]{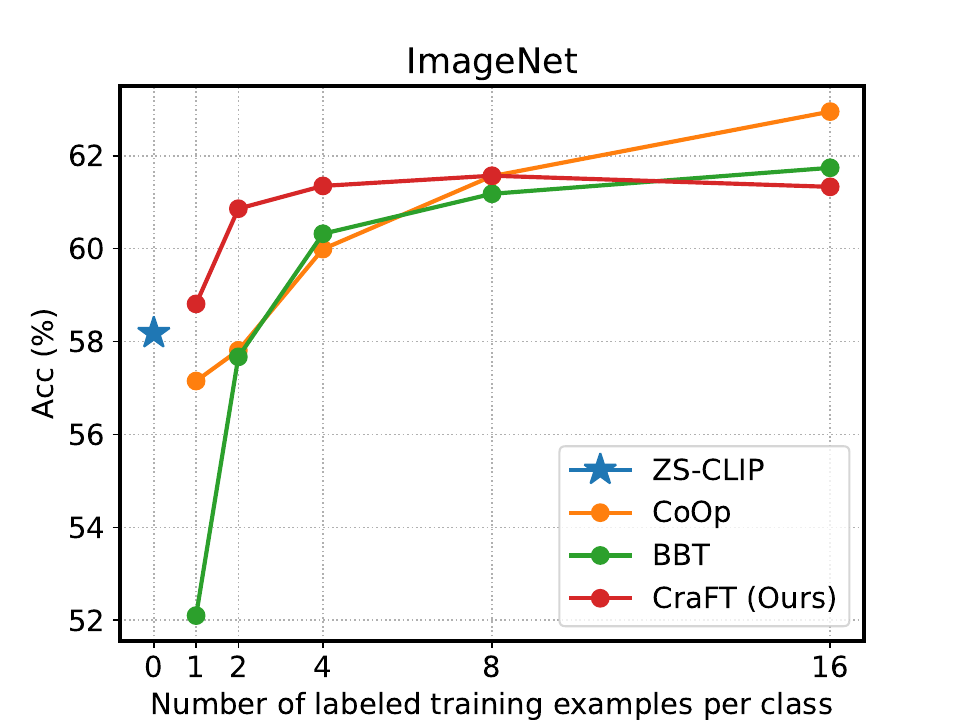}
    \end{minipage}
    \begin{minipage}{0.245\textwidth}
        \centering
        \includegraphics[width=1.13\textwidth]{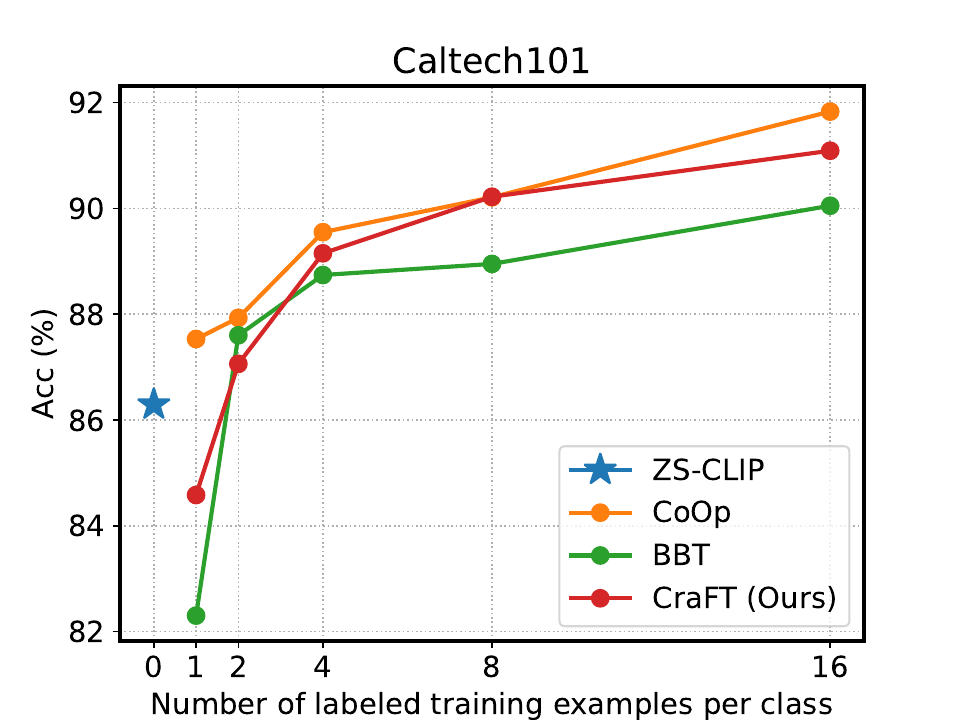}
    \end{minipage}
    \begin{minipage}{0.245\textwidth}
        \centering
        \includegraphics[width=1.13\textwidth]{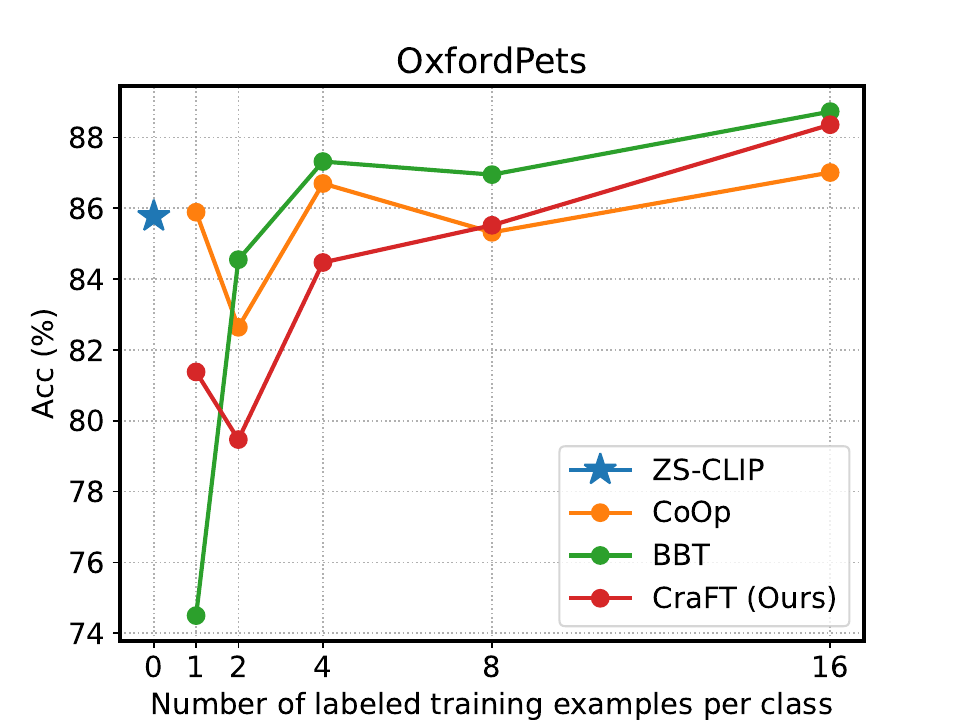}
    \end{minipage}
    \\
    \begin{minipage}{0.245\textwidth}
        \centering
        \includegraphics[width=1.13\textwidth]{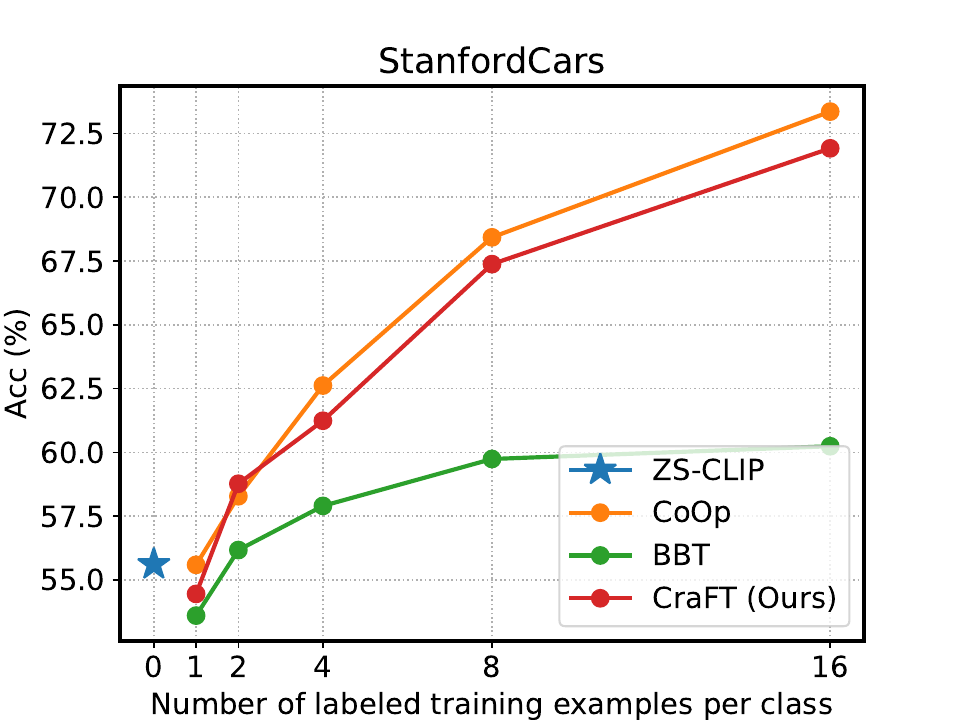}
    \end{minipage}
    \begin{minipage}{0.245\textwidth}
        \centering
        \includegraphics[width=1.13\textwidth]{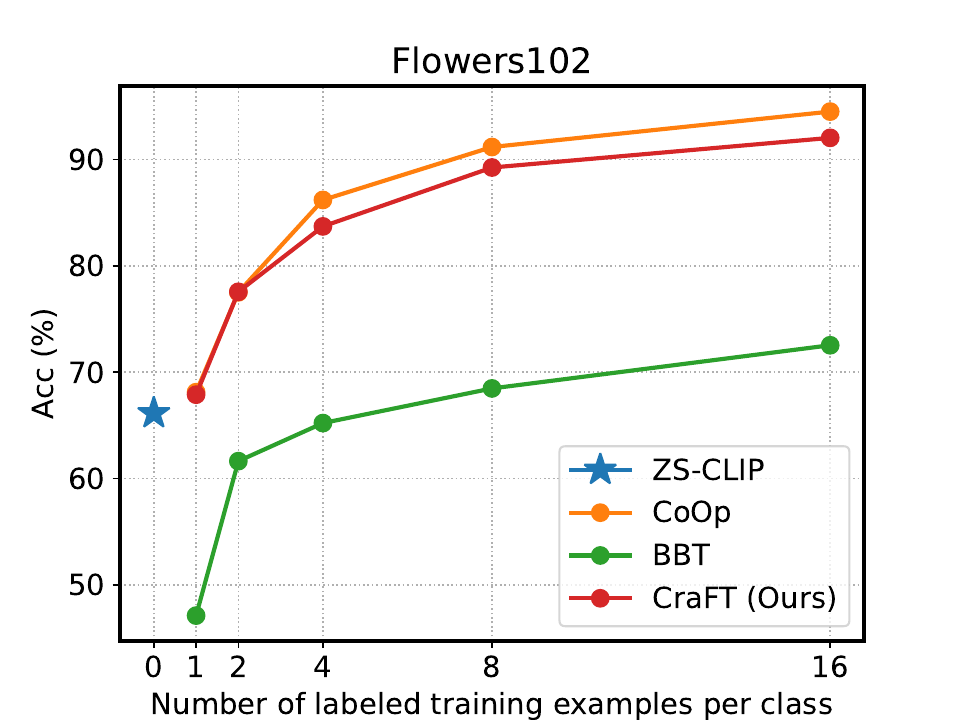}
    \end{minipage}
    \begin{minipage}{0.245\textwidth}
        \centering
        \includegraphics[width=1.13\textwidth]{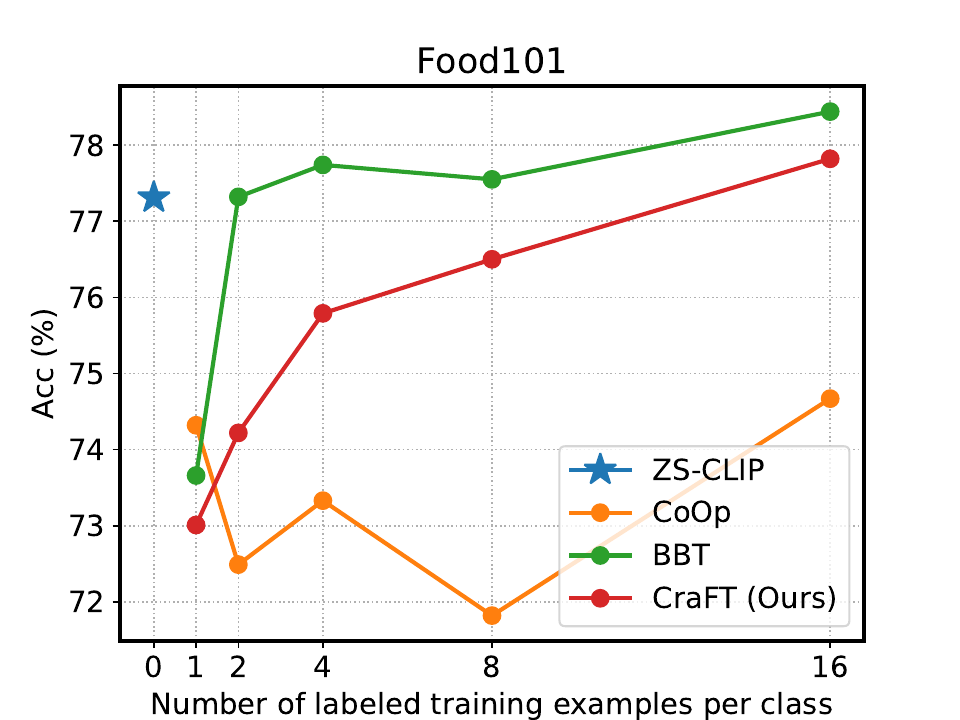}
    \end{minipage}
    \begin{minipage}{0.245\textwidth}
        \centering
        \includegraphics[width=1.13\textwidth]{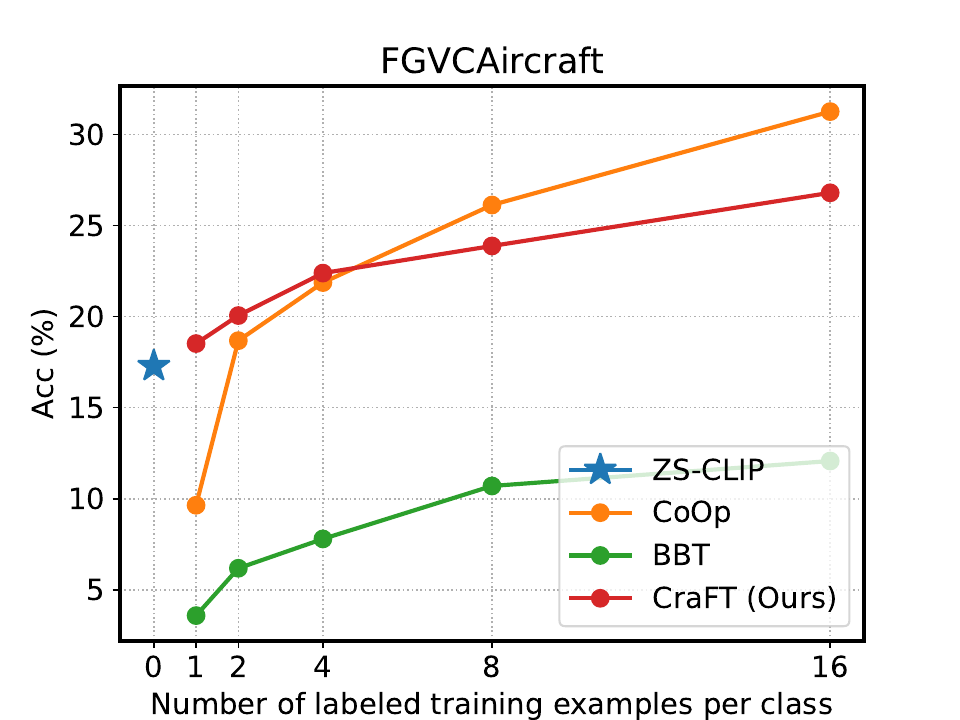}
    \end{minipage}
    \\
    \begin{minipage}{0.245\textwidth}
        \centering
        \includegraphics[width=1.13\textwidth]{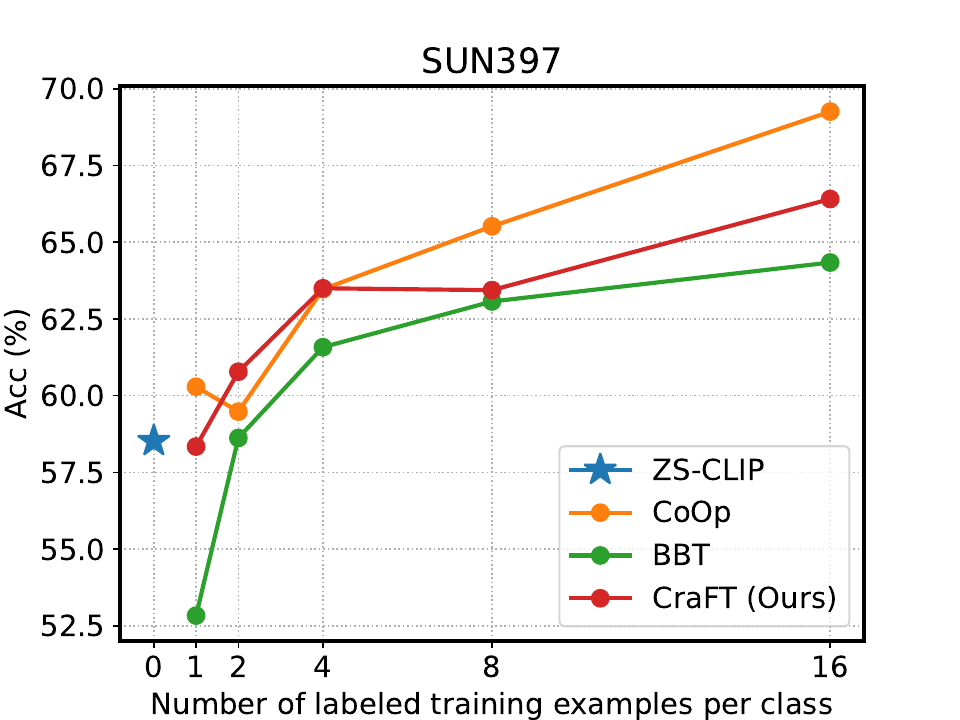}
    \end{minipage}
    \begin{minipage}{0.245\textwidth}
        \centering
        \includegraphics[width=1.13\textwidth]{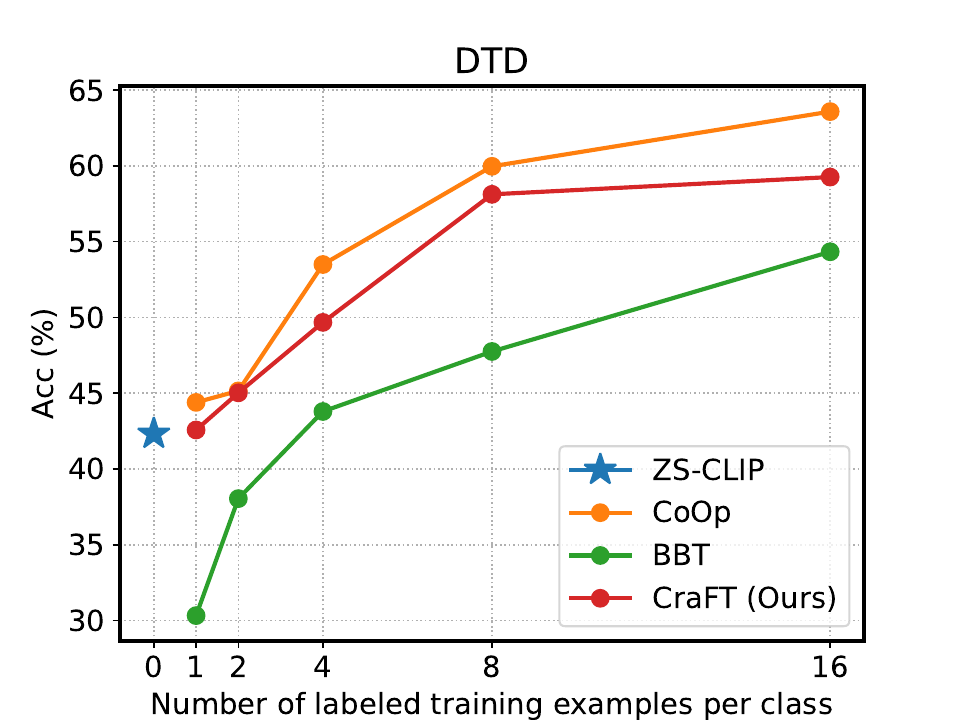}
    \end{minipage}
    \begin{minipage}{0.245\textwidth}
        \centering
        \includegraphics[width=1.13\textwidth]{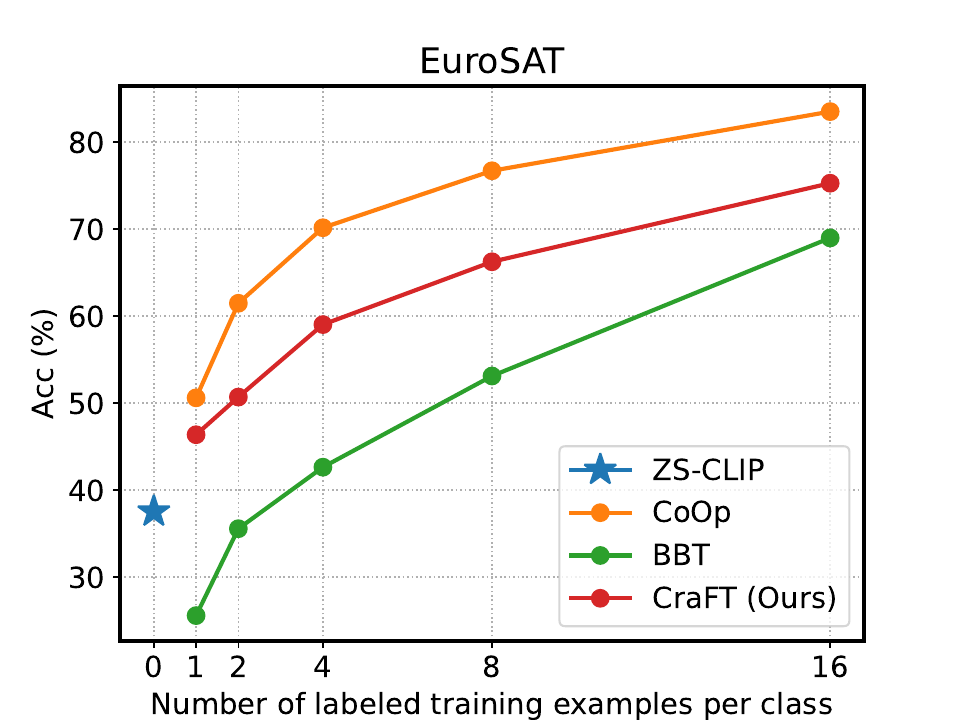}
    \end{minipage}
    \begin{minipage}{0.245\textwidth}
        \centering
        \includegraphics[width=1.13\textwidth]{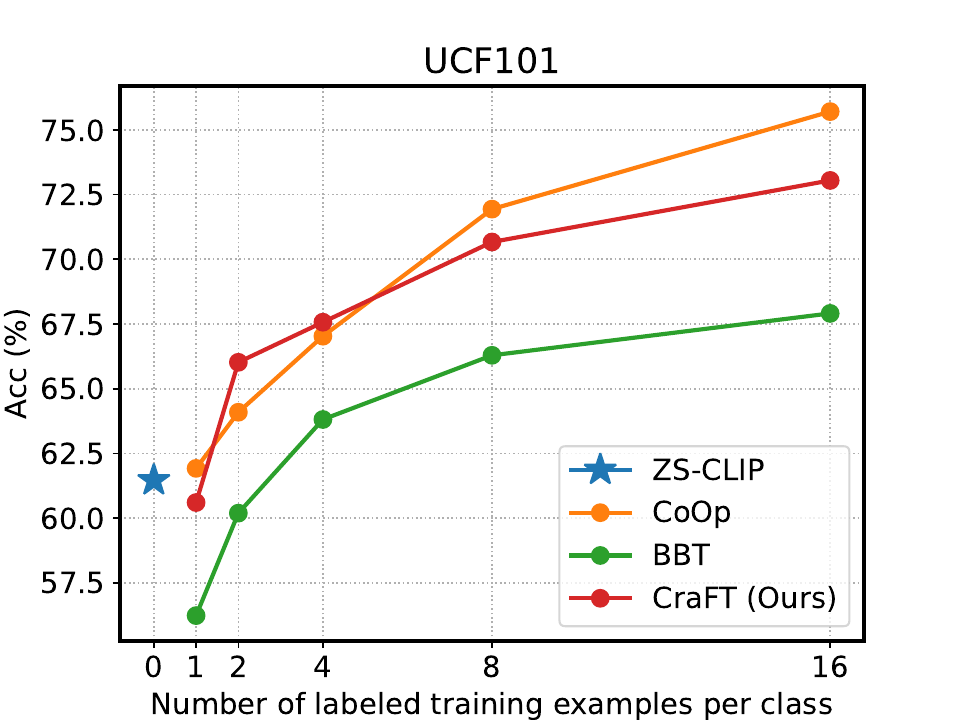}
    \end{minipage}\
    \vspace{-5pt}
    \caption{
    \textbf{Results of few-shot learning on the 11 datasets.}
    Here, CoOp is a white-box method that works as the upper bound. 
    Our proposed method greatly surpasses the black-box baseline methods, BBT, and zero-shot CLIP. 
   }
    \label{fig:fewshots}
    \vspace{-15pt}
\end{figure*}

\section{Experiments}
\label{sec:exp}

\subsection{Setup}
\textbf{Datasets.} 
In accordance with CoOp~\cite{zhou2022learning}, we adopt 11 distinct image classification datasets to investigate few-shot learning.
These datasets encompass various domains of image classification, including generic object recognition with ImageNet~\cite{deng2009imagenet} and Caltech101~\cite{fei2004learning}, fine-grained image recognition with OxfordPets~\cite{parkhi2012cats}, StanfordCars~\cite{krause20133d}, Flowers102~\cite{nilsback2008automated}, Food101~\cite{bossard2014food} and FGVCAircraft~\cite{maji2013fine}, satellite image classification with EuroSAT~\cite{helber2019eurosat}, action classification with UCF101~\cite{soomro2012ucf101}, texture classification with DTD~\cite{cimpoi2014describing}, and scene recognition with SUN397~\cite{xiao2010sun}.
Moreover, we utilize 4 additional datasets to investigate the robustness of the model to distribution, including ImageNetV2~\cite{recht2019imagenet}, ImageNet-Sketch~\cite{wang2019learning}, ImageNet-A~\cite{hendrycks2021natural}, and ImageNet-R~\cite{hendrycks2021many}.
ImageNetV2 is a reproduced test set using different sources while following ImageNet’s data collection process. ImageNet-Sketch contains sketch images belonging to the same 1,000 ImageNet classes. Both ImageNet-A and -R contains 200 classes derived from a subset of ImageNet’s 1,000 classes.

\noindent
\textbf{Evaluation Protocol.}
To evaluate the performance of few-shot learning models, we have followed the evaluation protocol proposed in CLIP~\cite{radford2021learning}. Specifically, we have trained models using 1, 2, 4, 8, and 16 shots and evaluated them on the full test sets. 
Additionally, we have assessed the robustness of the models to distribution shift by training CraFT on ImageNet~\cite{deng2009imagenet} with 16 shots and evaluating it on target datasets ImageNetV2~\cite{recht2019imagenet}, ImageNet-Sketch~\cite{wang2019learning}, ImageNet-A~\cite{hendrycks2021natural}, and ImageNet-R~\cite{hendrycks2021many}.

\setlength{\tabcolsep}{6pt}
\begin{table*}[!tbp]
  \centering
  \caption{
    \textbf{Results of different architectures on 11 datasets.} 
    The models are trained on the 16-shot setting datasets.
    \textbf{Bold} denotes the best results of black-box methods.
   }
   \vspace{-10pt}
  \resizebox{1.0\textwidth}{!}{
    \begin{tabular}{llcccccccccccc}
    \toprule
    \multicolumn{1}{l}{\textbf{Backbone}} &\multicolumn{1}{l}{\textbf{Method}} & \textbf{Pets} & \textbf{Flo} & \textbf{FGVC} & \textbf{DTD} & \textbf{Euro} & \textbf{Cars} & \textbf{Food} & \textbf{SUN} & \textbf{Cal} & \textbf{UCF} & \textbf{IN} & \textbf{Avg.} \\
    \midrule
    \multirow{3}{*}{ResNet-50} & \multicolumn{1}{l}{ZS-CLIP~\cite{radford2021learning}} & 85.77  & 66.14  & 17.28  & 42.32  & 37.56  & 55.61  & 77.31  & 58.52  & 86.29  & 61.46  & 58.18  & 58.77  \\
    & \multicolumn{1}{l}{BBT~\cite{sun2022black}} & \textbf{88.73}  & 72.53  & 12.07  & 54.33  & 69.01  & 60.24  & \textbf{78.44}  & 64.34  & 90.05  & 67.91  & \textbf{61.74}  & 65.40  \\
    \rowcolor{gray!40}  
    \cellcolor{white} & \multicolumn{1}{l}{CraFT} & 88.36  & \textbf{92.04}  & \textbf{26.80}  & \textbf{59.26}  & \textbf{75.30}  & \textbf{71.92}  & 77.82  & \textbf{66.41}  & \textbf{91.09}  & \textbf{73.05}  & 61.33  & \textbf{71.22}  \\
    \midrule
    \multirow{3}{*}{ResNet-101} & \multicolumn{1}{l}{ZS-CLIP~\cite{radford2021learning}} & 86.75  & 64.03  & 18.42  & 38.59  & 32.59  & 66.23  & 80.53  & 58.96  & 89.78  & 60.96  & 61.62  & 59.86  \\
    & \multicolumn{1}{l}{BBT~\cite{sun2022black}} & \textbf{89.44}  & 75.48  & 24.02  & 54.79  & 64.73  & 67.84  & \textbf{81.77}  & \textbf{65.52}  & 92.97  & 70.74  & \textbf{64.35}  & 68.33  \\
    \rowcolor{gray!40}  
    \cellcolor{white} & \multicolumn{1}{l}{CraFT} & 89.20  & \textbf{89.61}  & \textbf{28.05}  & \textbf{58.85}  & \textbf{67.16}  & \textbf{71.57}  & 81.37  & 64.79  & \textbf{93.21}  & \textbf{75.16}  & 63.68  & \textbf{71.15}  \\
    \midrule
    \multirow{3}{*}{ViT-B/32} & \multicolumn{1}{l}{ZS-CLIP~\cite{radford2021learning}} & 87.49  & 66.95  & 19.23  & 43.97  & 45.19  & 60.55  & 80.50  & 61.91  & 90.87  & 62.01  & 62.05  & 61.88  \\
    & \multicolumn{1}{l}{BBT~\cite{sun2022black}} & \textbf{89.77}  & 74.14  & 18.72  & 55.85  & 69.67  & 63.21  & \textbf{81.44}  & 68.08  & \textbf{94.20}  & 72.67  & \textbf{65.18}  & 68.45  \\
    \rowcolor{gray!40}  
    \cellcolor{white} & \multicolumn{1}{l}{CraFT} & 88.11  & \textbf{90.43}  & \textbf{26.24}  & \textbf{60.40}  & \textbf{70.07}  & \textbf{70.45}  & 77.89  & \textbf{68.57}  & 93.79  & \textbf{75.72}  & 62.76  & \textbf{71.31}  \\
    \midrule
    \multirow{3}{*}{ViT-B/16} & \multicolumn{1}{l}{ZS-CLIP~\cite{radford2021learning}} & 89.21  & 71.34  & 24.72  & 44.39  & 47.60  & 65.32  & 86.06  & 62.50  & 92.94  & 66.75  & 66.73  & 65.23  \\
    & \multicolumn{1}{l}{BBT~\cite{sun2022black}} & \textbf{92.70}  & 82.41  & 29.49  & 59.26  & 70.48  & 70.19  & \textbf{86.42}  & 70.33  & \textbf{94.75}  & 70.48  & \textbf{70.15}  & 72.42  \\
    \rowcolor{gray!40}  
    \cellcolor{white} & \multicolumn{1}{l}{CraFT} & 91.94  & \textbf{93.92}  & \textbf{36.89}  & \textbf{63.28}  & \textbf{72.07}  & \textbf{78.11}  & 83.66  & \textbf{70.97}  & 94.48  & \textbf{79.78}  & 68.21  & \textbf{75.76}  \\
    \bottomrule
    \end{tabular}
  }
  \label{tab:architecture}%
  \vspace{-15pt}
\end{table*}%

\noindent
\textbf{Training Details.}
We utilize CLIP~\cite{radford2021learning} as our black-box vision-language model, with ResNet-50~\cite{he2016deep} and transformer~\cite{vaswani2017attention} serving as the vision and language encoders, respectively.
These encoders are initialized with CLIP's pretrained weights and kept frozen and unseen during training.
To optimize the text prompts in the prompt generation module, we used the CMA-ES algorithm and set the prompt length to 4.
The text prompts are projected into a subspace of dimension 512 using a random matrix sampled from a Gaussian distribution $\mathcal{N}(0, 0.02)$. 
The population size is set to 40, with a budget of 8,000 API calls.
For the prediction refinement module, we use a three-layer MLP with a hidden dimension of 512 as the refinement network.
We set the hyper-parameters $\lambda_I$ and $\lambda_O$ to 0.1 divided by the number of classes by default.
The prediction refinement module is optimized using the AdamW optimizer with a learning rate of 0.001, and we set the batch size as 256 during training. 
Results are reported with average accuracy.
All experiments are conducted on a single NVIDIA GeForce RTX 3090.
We conducted three runs with different seeds and averaged the results to obtain a reliable estimate of model performance.

\noindent
\textbf{Baseline Methods.}
To evaluate the effectiveness of CraFT, we compare it with three baseline methods.
\textbf{(1) ZS-CLIP:} 
Our first baseline method is zero-shot CLIP~\cite{radford2021learning}. This method requires handcrafted prompts, which we set to be the same as those used in previous works~\cite{zhou2022learning, zhou2022conditional} to ensure a fair comparison.
\textbf{(2) CoOp:}
Our second baseline method is CoOp~\cite{zhou2022learning}. CoOp is a white-box method that proposes learning the global text prompts through gradient descent. We use the best version of CoOp~\cite{zhou2022learning}, setting the length of text prompts to 16, for comparison.
\textbf{(3) BBT:} 
Our third baseline method is Black-Box Tuning (BBT)~\cite{sun2022black}. BBT is a black-box method for NLP tasks that proposes optimizing the soft prompts with the CMA-ES algorithm. We implement BBT in the black-box VLM, and we set its hyper-parameters the same as CraFT.

\subsection{Results of Few-Shot Classification}
Figure~\ref{fig:fewshots} illustrates the performance of our proposed method, CraFT, in comparison to three baseline methods: CoOp~\cite{zhou2022learning}, BBT~\cite{sun2022black}, and ZS-CLIP~\cite{radford2021learning}, across 11 downstream datasets, accompanied by their respective average results.
Our proposed approach demonstrates a significant superiority over the other black-box methods, i.e., ZS-CLIP and BBT.
Specifically, under the 16-shot setting, CraFT achieves a substantial accuracy improvement of 12.45\% and 5.82\% when compared to ZS-CLIP and BBT, respectively.

Our proposed CraFT surpasses the black-box baseline BBT~\cite{sun2022black} on most datasets except OxfordPets~\cite{parkhi2012cats} and Food101~\cite{bossard2014food}.
OxfordPets~\cite{parkhi2012cats} and Food101~\cite{bossard2014food} are fine-grained datasets and therefore sensitive to the fine-tuning process.
Since gradient-based optimization estimates the gradient on batch inputs, it can lead to unstable training and hurt the good properties of the pretrained model.
Therefore, on OxfordPets~\cite{parkhi2012cats} and Food101~\cite{bossard2014food} datasets, BBT, which is optimized without gradient, performs significantly better than gradient-based methods (CoOp and our CraFT).

\subsection{Effectiveness of Different Architectures}
We further evaluate the effectiveness of our proposed method on the 11 datasets with different visual architectures of CLIP~\cite{radford2021learning}, containing both CNNs~\cite{he2016deep} and ViTs~\cite{dosovitskiy2020image}.
Table~\ref{tab:architecture} shows the results of our methods and two black-box baselines, ZS-CLIP and BBT, with different model architectures.
These methods are trained on downstream 16-shot datasets.

On average, our proposed CraFT method outperformed zero-shot CLIP~\cite{radford2021learning} by 12.45\%, 11.29\%, 9.43\%, and 10.53\% on ResNet-50, ResNet-101, ViT-B/32, and ViT-B/16-based CLIP, respectively.
Additionally, CraFT outperforms BBT~\cite{sun2022black} by 5.82\%, 2.82\%, 2.86\%, and 3.34\% on average of the 11 datasets on ResNet-50, ResNet-101, ViT-B/32, and ViT-B/16 based CLIP, respectively.
These results demonstrate the effectiveness of CraFT across different black-box model architectures.

\setlength{\tabcolsep}{4pt}
\begin{table*}[!tbp]
  \centering
  \caption{
    \textbf{Robustness to distribution shift.}
    We compare our method with CLIP and CoOp (prompt length $L=4$ and $L=16$).
    And the models are trained on 16-shot datasets with different architectures.
    \textbf{Bold} and \underline{Underline} denote the highest and second highest results.
 }
 \vspace{-10pt}
  \resizebox{0.75\textwidth}{!}{
    \begin{tabular}{lllccccccc}
    \toprule
    \multicolumn{1}{l}{\multirow{2}[4]{*}{Backbone}} & \multicolumn{2}{l}{\multirow{2}[4]{*}{Method}} & \multicolumn{1}{c}{\multirow{2}[4]{*}{Black-Box}} & Source & \multicolumn{5}{c}{Target} \\
    \cmidrule(r){5-5} \cmidrule(r){6-10} 
    \multicolumn{3}{c}{} &       & ImageNet & -V2   & -Sketch & -A    & -R    & Avg. \\
    \midrule
    \multirow{4}{*}{ResNet-50} &
    \multicolumn{2}{l}{ZS-CLIP~\cite{radford2021learning}} & \ding{51}  & 58.18  & 51.34  & 33.32  & 21.65  & 56.00  & 40.58  \\
    & \multicolumn{2}{l}{CoOp (L=4)~\cite{zhou2022learning}} & \ding{55} & \textbf{63.33} & \textbf{55.40} & \textbf{34.67} & \underline{23.06}  & \underline{56.60}  & \textbf{42.43} \\
    & \multicolumn{2}{l}{CoOp (L=16)~\cite{zhou2022learning}} & \ding{55} & \underline{62.95}  & \underline{55.11}  & 32.74  & 22.12  & 54.96  & 41.23  \\
    \rowcolor{gray!40}  
    \cellcolor{white}& \multicolumn{2}{l}{CraFT} & \ding{51}  & 61.33   & 53.92  & \underline{34.01}  & \textbf{23.13} & \textbf{58.23} & \underline{42.32}  \\
    \midrule
    \multirow{4}{*}{ResNet-101} &
    \multicolumn{2}{l}{ZS-CLIP~\cite{radford2021learning}} & \ding{51}  & 61.62  & 54.81  & 38.71  & 28.05  & 64.38  & 46.49  \\
    & \multicolumn{2}{l}{CoOp (L=4)~\cite{zhou2022learning}} & \ding{55} & \underline{65.98}  & \underline{58.60}  & \textbf{40.40} & \underline{29.60}  & \underline{64.98}  & \textbf{48.40} \\
    & \multicolumn{2}{l}{CoOp (L=16)~\cite{zhou2022learning}} & \ding{55} & \textbf{66.60} & \textbf{58.66} & 39.08  & 28.89  & 63.00  & 47.41  \\
    \rowcolor{gray!40}  
    \cellcolor{white}& \multicolumn{2}{l}{CraFT} & \ding{51}  & 63.68  & 56.95  & \underline{39.50}  & \textbf{30.40} & \textbf{65.94} & \underline{48.20}  \\
    \midrule
    \multirow{4}{*}{ViT-B/32} &
    \multicolumn{2}{l}{ZS-CLIP~\cite{radford2021learning}} & \ding{51}  & 62.05  & 54.79  & 40.82  & 29.57  & \underline{65.99}  & 47.79  \\
    & \multicolumn{2}{l}{CoOp (L=4)~\cite{zhou2022learning}} & \ding{55} & \underline{66.34}  & \textbf{58.24} & \textbf{41.48} & \textbf{31.34} & 65.78  & \textbf{49.21} \\
    & \multicolumn{2}{l}{CoOp (L=16)~\cite{zhou2022learning}} & \ding{55} & \textbf{66.85} & \underline{58.08}  & \underline{40.44}  & 30.62  & 64.45  & 48.40  \\
    \rowcolor{gray!40}  
    \cellcolor{white} & \multicolumn{2}{l}{CraFT} & \ding{51}  & 62.76  & 56.55  & 40.26  & \underline{31.27}  & \textbf{66.46} & \underline{48.63}  \\
    \midrule
    \multirow{4}{*}{ViT-B/16} &
    \multicolumn{2}{l}{ZS-CLIP~\cite{radford2021learning}} & \ding{51}  & 66.73  & 60.83  & 46.15  & 47.77  & 73.96  & 57.18  \\
    & \multicolumn{2}{l}{CoOp (L=4)~\cite{zhou2022learning}} & \ding{55} & \underline{71.73}  & \textbf{64.56} & \textbf{47.89} & \textbf{49.93} & 75.14  & \textbf{59.38} \\
    & \multicolumn{2}{l}{CoOp (L=16)~\cite{zhou2022learning}} & \ding{55} & \textbf{71.92} & \underline{64.18}  & 46.71  & 48.41  & 74.32  & 58.41  \\
    \rowcolor{gray!40}  
    \cellcolor{white} & \multicolumn{2}{l}{CraFT} & \ding{51}  & 68.21  & 62.78  & \underline{47.17}  & \underline{49.73}  & \textbf{75.52} & \underline{58.80}  \\
    \bottomrule
    \end{tabular}%
   }
  \label{tab:robustness}%
  \vspace{-15pt}
\end{table*}%

\subsection{Robustness to Distribution Shift}
We further conduct experiments to evaluate the robustness of CraFT to distribution shift. 
Specifically, we trained the models using the 16-shot ImageNet~\cite{deng2009imagenet} dataset and subsequently transferred them to target domain shift datasets.
These included ImageNetV2~\cite{recht2019imagenet}, ImageNet-Sketch~\cite{wang2019learning}, ImageNet-A~\cite{hendrycks2021natural}, and ImageNet-R~\cite{hendrycks2021many}.

Table~\ref{tab:robustness} reports the results of our method and two other baseline methods: ZS-CLIP~\cite{radford2021learning} and CoOp~\cite{zhou2022learning}  (prompt length $L=4$ and $L=16$).
Our proposed CraFT outperforms ZS-CLIP on all datasets and architectures, with improvements of 1.74\%, 1.71\%, 0.84\%, and 16.2\% observed for ResNet-50, ResNet-101, ViT-B/32, and ViT-B/16-based CLIP, respectively.
These results illustrate that CraFT enhances the robustness of CLIP~\cite{radford2021learning}.
Moreover, our method CraFT achieves comparable performance with the white-box prompt tuning method, CoOp.
Compared with the CoOp ($L=16$) variant, which performs well on few-shot classification, our CraFT achieves improvements of 1.09\%, 0.79\%, 0.23\%, and 0.39\% for each architecture, yielding the effectiveness of our method.

\subsection{Ablation Study}

\textbf{Effectiveness of Components.}
In this part, we analyze the efficacy of the components of CraFT. 
Table~\ref{tab:effectiveness} displays the average results obtained from 11 downstream datasets for different shot settings.
In the table, ``PG." refers to the prompt generation module, ``PR." indicates the prediction refinement module, and ``Co." stands for the collaborative training algorithm.

The results demonstrate that using either the prompt generation module or the prediction refinement module in isolation achieves superior performance compared to ZS-CLIP (58.77\%) in downstream tasks.
This indicates the effectiveness of both the prompt generation module and the prediction refinement module.
However, when optimizing them iteratively without using the collaborative training algorithm, the model performs even worse than using the prediction refinement module alone.
After incorporating the collaborative training algorithm, the models exhibit better performance compared to the other settings, indicating the effectiveness of this component.
Therefore, it can be concluded that both the prompt generation module and prediction refinement module are effective, and they work best when optimized together using the collaborative training.
\setlength{\tabcolsep}{6pt}
\begin{table}[tbp]
    \captionof{table}{
        We ablate the components of CraFT with different shots.
        PG. denotes the prompt generation module.
        PR. denotes the prediction refinement module.
        Co. denotes the collaborative training algorithm.
        Results are average over 11 dataset.
   }  
    \label{tab:effectiveness}
    \resizebox{0.45\textwidth}{!}{
    \begin{tabular}{ccc|ccccc}
    \toprule
    PG.   & PR. & Co.  & 1     & 2     & 4     & 8     & 16 \\
    \midrule
    \ding{51}     &   \ding{55}    &    \ding{55}   & 50.17  & 56.69  & 59.71  & 62.16  & 65.40  \\
    \ding{55}    & \ding{51}     &    \ding{55}   & 54.54  & 59.07  & 63.13  & 66.07  & 69.29  \\
    \ding{51}     & \ding{51}     &   \ding{55}    & 51.54  & 56.93  & 60.70  & 64.47  & 68.23  \\
    \ding{51}     & \ding{51}     & \ding{51}     & \textbf{59.49}  & \textbf{61.87}  & \textbf{65.26}  & \textbf{68.44}  & \textbf{71.22}  \\
    \bottomrule
    \end{tabular}%
   }
\end{table}

\setlength{\tabcolsep}{6pt}
\begin{table*}[tbp]
  \centering
  \caption{
  Comparison of deployment efficiency, the viability of black-box, test accuracy, training time, and memory footprint of user and server. 
  Models are trained using RN50 CLIP with the 16-shot ImageNet.
 }
  \resizebox{0.8\textwidth}{!}{
    \begin{tabular}{cc|cccccccc}
    \toprule
    \multicolumn{2}{c|}{Method} & \multicolumn{2}{c}{Black-Box} & \multicolumn{2}{c}{Test Accuracy} & \multicolumn{2}{c}{Training Time} & Mem. (User) & Mem. (Server) \\
    \midrule
    \multicolumn{2}{l|}{ZS-CLIP~\cite{radford2021learning}} & \multicolumn{2}{c}{\ding{51}} & \multicolumn{2}{c}{58.18} & \multicolumn{2}{c}{0} & 0  &  244.7 MB \\
    \multicolumn{2}{l|}{CoOp~\cite{zhou2022learning}} & \multicolumn{2}{c}{\ding{55}} & \multicolumn{2}{c}{62.95} & \multicolumn{2}{c}{ 2h 3min} &    395.7 MB   &  0 \\
    \rowcolor{gray!40}  
    \multicolumn{2}{l|}{CraFT} & \multicolumn{2}{c}{\ding{51}} & \multicolumn{2}{c}{61.33} & \multicolumn{2}{c}{1h 44min} &    5.0 MB   & 244.7 MB  \\
    \bottomrule
    \end{tabular}%
   }
  \label{tab:deployment}%
\end{table*}%

\textbf{Ablation of the Prediction Refinement Network.}
Furthermore, we investigate the best architecture of the prediction refinement module. In Table~\ref{tab:refinement}, we ablate the effectiveness of the residual connection and the architecture of the refinement network $R$.
As shown in Table~\ref{tab:refinement}, the performance of CraFT drops dramatically if we delete the residual connection (-30.91\% on the 1-shot setting), indicating its effectiveness.
Additionally, changing the MLP refinement network to a linear mapping also results in a significant performance drop.
As a result, we implement the prediction refinement module using the MLP as the refinement network together with a shortcut connection.  
\setlength{\tabcolsep}{6pt}
\begin{table}[tbp]
  \centering
  \caption{
    \textbf{Ablation of hyper-parameter $\lambda_I$}. 
    Models are trained using RN50 CLIP in EuroSAT.
  }
  \resizebox{0.45\textwidth}{!}{
    \begin{tabular}{cc|cccccc}
    \toprule
    \multicolumn{2}{c|}{$\lambda_I$} & 0.00     & 0.01  & 0.03  & 0.05  & 0.07  & 0.09 \\
    \midrule
    \multicolumn{2}{c|}{CraFT} & 72.51  & 75.30  & 73.83  & 76.86  & 72.83  & 70.72  \\
    \bottomrule
    \end{tabular}%
    }
    \vspace{-10pt}
  \label{tab:ablation_i}%
\end{table}%
\setlength{\tabcolsep}{6pt}
\begin{table}[t]
  \centering
  \caption{
    \textbf{Ablation of hyper-parameter $\lambda_O$}. 
    Models are trained using RN50 CLIP in EuroSAT.
  }
  \resizebox{0.45\textwidth}{!}{
    \begin{tabular}{cc|cccccc}
    \toprule
    \multicolumn{2}{c|}{$\lambda_O$} & 0.00     & 0.01  & 0.03  & 0.05  & 0.07  & 0.09 \\
    \midrule
    \multicolumn{2}{c|}{CraFT} & 74.45  & 75.30  & 74.92  & 75.91  & 74.83  & 75.49  \\
    \bottomrule
    \end{tabular}%
    }
  \label{tab:ablation_o}%
\end{table}%
\setlength{\tabcolsep}{6pt}
\begin{table}[t]
    \centering
    \captionof{table}{
    We ablate the components of the prediction refinement module.
    Arch. denotes the architecture of the refinement network.
    \textbf{Bold} denotes the highest result.
   }
    \label{tab:refinement}%
    \resizebox{0.45\textwidth}{!}{   
    \begin{tabular}{cc|ccccc}
    \toprule
          &       & \multicolumn{5}{c}{shots} \\
    Residual & Arch.     & 1     & 2     & 4     & 8     & 16 \\
    \midrule
    \ding{55}  & MLP   & 28.58  & 44.44  & 52.58  & 59.25  & 63.10  \\
    \ding{51}  & Linear & 47.99  & 52.70  & 56.74  & 62.84  & 65.88  \\
    \ding{51}  & MLP   & \textbf{59.49}  & \textbf{61.87}  & \textbf{65.26}  & \textbf{68.44}  & \textbf{71.22}  \\
    \bottomrule
    \end{tabular}%
   }
\end{table}

\textbf{Ablation of Hyper-parameters.}
Moreover, we ablate the sensitivity of the hyper-parameters $\lambda_I$ and $\lambda_O$.
The default values for $\lambda_I$ and $\lambda_O$ are set to 0.1 divided by the number of classes, which is 0.01 in this case.
As shown in Table~\ref{tab:ablation_i} and Table~\ref{tab:ablation_o}, the model performance does not exhibit a significant correlation with the values of $\lambda_I$ and $\lambda_O$.
Nevertheless, it consistently outperforms the case where lambda is set to 0, indicating the effectiveness of the consistency loss.
Consequently, we maintain the default values of $\lambda_I$ and $\lambda_O$ across all experimental settings and datasets.

\textbf{Comparison of Efficiency.}
We further evaluate the efficiency of CraFT and compare it with CoOp~\cite{zhou2022learning} and ZS-CLIP~\cite{radford2021learning} on the 16-shot ImageNet dataset, based on deployment efficiency, black-box viability, test accuracy, training time, and memory footprint.
Table~\ref{tab:deployment} shows that CraFT trains faster and has a significantly smaller memory footprint, using only 1/80 of the memory footprint of the white-box method CoOp, and CraFT incurs only a marginal loss in test accuracy.

\begin{figure}
    \centering
    \includegraphics[trim={0.15cm 0.15cm 0.15cm 0.15cm},clip,width=0.375\textwidth]{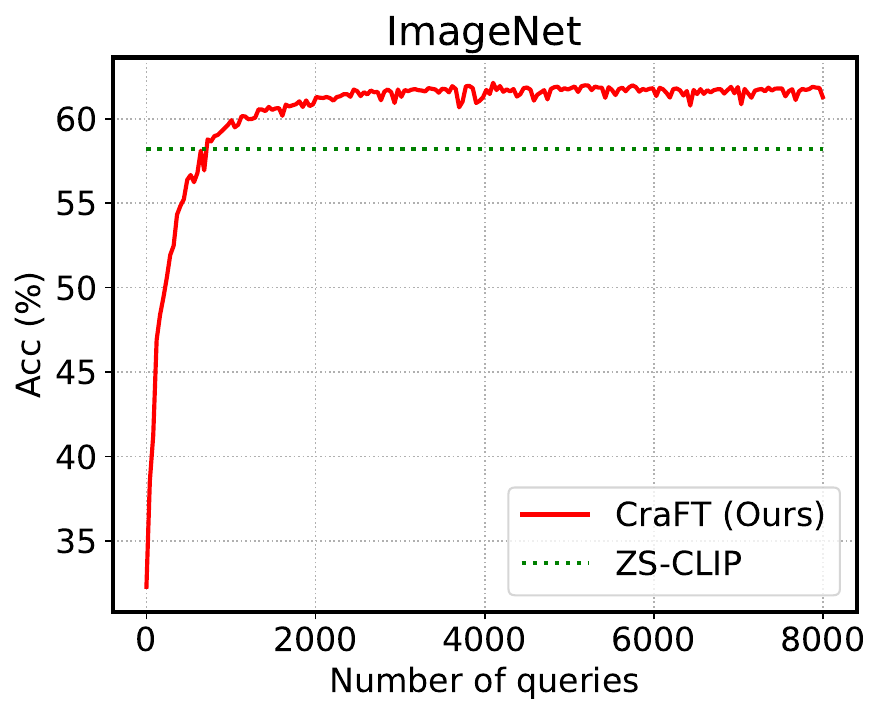}
    \caption{
        The relation of performance and number of queries on 16-shot ImageNet.
        The models utilize RN50 CLIP.
   }
    \label{fig:query}   
\end{figure}

We further report the query efficacy of our method.
For the black-box vision-language model, we first upload the image dataset to the server, and we query the model with the text prompts to obtain its output $f(\{t_i\}_{i=1}^K, \{i_j\}_{j=1}^N)\in\mathbb{R}^{N\times K}$.
Figure~\ref{fig:query} depicts the relationship between the model's performance and the number of queries on the ImageNet dataset.
The results indicate that our method achieves a high degree of efficacy in terms of query efficiency.
Our method outperforms the ZS-CLIP approach with a mere 1,000 queries.
These results suggest its potential usefulness in a range of applications involving black-box vision-language models.

\section{Conclusion}
\label{sec:conclusion}
In this paper, we present \textbf{C}ollabo\textbf{ra}tive \textbf{F}ine-\textbf{T}uning (\textbf{CraFT}), a novel approach for fine-tuning black-box VLMs.
CraFT comprises a prompt generation module and a prediction refinement module, designed to learn the text prompts and refine the black-box output prediction, respectively.
Moreover, we developed a novel collaborative training algorithm capable of optimizing both modules jointly and mitigating conflicts between them.
We demonstrate the effectiveness of CraFT over 15 datasets, as well as its robustness to distribution shifts and different architectures.
Moreover, without the need for access to the parameters of vision-language models, CraFT improves its performance with marginal deployment cost and training costs.
These results demonstrate the effectiveness of our method.

\section*{Acknowledgements}
This work was funded by the Beijing Nova Program under Grant Z211100002121108, the National Natural Science Foundation of China under Grant 62276256, and the Young Elite Scientists Sponsorship Program by CAST (2023QNRC001).

\section*{Impact Statement}
This paper presents work whose goal is to advance the field of Machine Learning. There are many potential societal consequences of our work, none which we feel must be specifically highlighted here.

\nocite{langley00}

\bibliography{example_paper}
\bibliographystyle{icml2024}

\newpage
\appendix
\onecolumn

\section{Comparison of Black-Box Setting in VLMs}
\label{appendix:setting}
\setlength{\tabcolsep}{6pt}
\begin{table}[htbp]
  \centering
  \caption{
  \textbf{Comparision of black-box settings.}
  We compare the black-box assumption in this paper with assumptions in other methods in three aspects: VLM requirements, Network requirements, and Access Permission.
  }
  \resizebox{0.8\textwidth}{!}{
    \begin{tabular}{cc|cc|cc|c}
    \toprule
    \multicolumn{2}{c|}{\multirow{2}[3]{*}{Method}} & \multicolumn{5}{c}{Black-Box Assumption} \\
\cmidrule{3-7}    \multicolumn{2}{c|}{} & \multicolumn{2}{c|}{VLM requirements} & \multicolumn{2}{c|}{Network Requirements} & Access Permission \\
    \midrule    
    \multicolumn{2}{l|}{LFA~\cite{ouali2023black}} & \multicolumn{2}{c|}{\textcolor[RGB]{185, 37, 49}{\textbf{Dual-towered}}} & \multicolumn{2}{c|}{\textcolor[RGB]{0, 155, 158}{\textbf{No requirements}}} & \textcolor[RGB]{185, 37, 49}{\textbf{Features}} \\
    \multicolumn{2}{l|}{CBBT~\cite{guo2023black}} & \multicolumn{2}{c|}{\textcolor[RGB]{185, 37, 49}{\textbf{Dual-towered}}} & \multicolumn{2}{c|}{\textcolor[RGB]{0, 155, 158}{\textbf{No requirements}}} & \textcolor[RGB]{185, 37, 49}{\textbf{Features}} \\
    \multicolumn{2}{l|}{\citet{yu2023black}} & \multicolumn{2}{c|}{\textcolor[RGB]{0, 155, 158}{\textbf{No requirements}}} & \multicolumn{2}{c|}{\textcolor[RGB]{185, 37, 49}{\textbf{Vision Transformer}}} & \textcolor[RGB]{0, 155, 158}{\textbf{Predictions}} \\
    \midrule
    \multicolumn{2}{l|}{CraFT (Ours)} & \multicolumn{2}{c|}{\textcolor[RGB]{0, 155, 158}{\textbf{No requirements}}} & \multicolumn{2}{c|}{\textcolor[RGB]{0, 155, 158}{\textbf{No requirements}}} & \textcolor[RGB]{0, 155, 158}{\textbf{Predictions}} \\
    \bottomrule
    \end{tabular}%
    }
  \label{tab:comparison}%
\end{table}%

To illustrate the validity of our black-box assumption, we compared our black-box assumption with black-box assumptions in previous methods.
As shown in Table~\ref{tab:comparison}, we are comparing them in terms of VLM requirements, Network requirements, and Access Permission.

In comparison to LFA~\cite{ouali2023black} and CBBT~\cite{guo2023black}, our approach does not impose specific requirements on VLM, while LFA and CBBT need it to be dual-towered.
Notably, some VLMs are not dual-towered, and they have complex interactions between language and vision in the model, such as ALBEF~\cite{li2021align} and KOSMOS-1~\cite{huang2023language}. 
The setup in these methods may not be effective for these models.
Moreover, these methods require the black-box VLM to return features of each modality, potentially compromising model ownership.   
Granting such access to users poses a significant risk, as it opens the door for model theft through distillation.
Therefore, many black-box methods~\cite{yang2022divide, liang2022dine} opt for returning predictions instead of features.

In comparison to~\citet{yu2023black}, a crucial assumption in their work is that the vision encoder employed is a vision transformer.
This enables the incorporation of learnable prompts in the visual encoder, and the prompts can be updated through CMA-ES.
However, it is worth noting that many vision encoders in the realm of VLM are not based on the ViT architecture, such as the ResNet-based CLIP. 
Consequently, this assumption imposes constraints on the applicability of their methodology.
Furthermore, in Table~\ref{tab:more_experiemnt} presented below, a comparison between our CraFT and \citet{yu2023black} in the same architecture reveals that our method outperforms theirs, underscoring the efficacy of our approach.

In summary, we believe our black-box definition for VLM is more accurate and applicable to a wider range of scenarios.

\section{More Experimental Results}
To further verify the effectiveness of our method, we compare our method with two more black-box optimization methods, BlackVIP~\cite{oh2023blackvip} and \citet{yu2023black}.
We do not include the results in the main text due to the time-intensive nature of training these methods.
Additionally, the black-box framework assumed in \citet{yu2023black} is not consistent with ours.
Their framework presupposes access to the visual architecture of VLMs, assuming it to be a vision transformer. In contrast, we assume the model is invisible, and no assumptions are made about its architecture in our setting.
We believe our black-box setting is more reasonable and has broader applicability.

The comparison results are shown in Table~\ref{tab:more_experiemnt}.
We trained these methods with ViT-B/16 CLIP on 16-shot datasets.
To align with \citet{yu2023black}, we train a variant of our method where we also incorporate visual prompts (vp) in the black-box optimization.
As shown in the table, our method achieves the best results on average over the 11 datasets, even without including the vision prompts.
It surpasses BBT~\cite{sun2022black}, BlackVIP~\cite{oh2023blackvip}, and \citet{yu2023black} by 3.34\%, 7.54\%, and 2.04\%.  
Moreover, after incorporating the vision prompts in optimization, our method achieves an additional gain of 2.64\%, demonstrating that our method is compatible with visual prompts.

Moreover, we report the training time of these methods and our method in Table~\ref{tab:time_cost}.
The results show that our method is quite efficient compared with the baselines, which takes only about 2 days for training but achieves the best performance.
Moreover, we notice that including optimizing visual prompts will greatly increase the training time.
There may be two reasons contributing to this issue. 
Firstly, it increases the search dimensions of CMA-ES, which consequently decelerates the optimization algorithm. 
Secondly, for each image, multiple insertions of visual prompts are required (i.e., different sampled solutions in CMA-ES), followed by forward computations to obtain visual representations. 
In contrast, when visual prompts are not optimized, a single forward pass is sufficient to obtain the image representation in the whole training process.

In summary, the results show that our method is effective and efficient, and can be extended with visual prompts.

\setlength{\tabcolsep}{6pt}
\begin{table}[htbp]
  \centering
  \caption{
    \textbf{Comparing with more black-box methods.}
    The methods are trained with ViT-B/16 CLIP on 16-shot datasets.
    `vp' denotes that we include the same visual prompts for optimization as \citet{yu2023black} for a fair comparison. 
    }
\resizebox{1.0\textwidth}{!}{
\begin{tabular}{l|ccccccccccccc}
\toprule   
\textbf{Method} & \textbf{Pets} & \textbf{Flo} & \textbf{FGVC} & \textbf{DTD} & \textbf{Euro} & \textbf{Cars} & \textbf{Food} & \textbf{SUN} & \textbf{Cal} & \textbf{UCF} & \textbf{IN} & \textbf{Avg.} \\
\midrule
    \multicolumn{1}{l|}{BBT} & \underline{92.70}  & 82.41  & 29.49  & 59.26  & 70.48  & 70.19  & 86.42  & 70.33  & \underline{94.75}  & 70.48  & \textbf{70.15}  & 72.42  \\
    \multicolumn{1}{l|}{BlackVIP} & 89.70  & 70.60  & 25.00  & 45.20  & 73.10  & 65.60  & \underline{86.60}  & 64.70  & 93.70  & 69.10  & 67.10  & 68.22  \\
    \multicolumn{1}{l|}{\citet{yu2023black}} & 93.43  & 82.87  & 30.08  & 59.24  & \underline{82.33}  & 68.39  & \textbf{86.94}  & 69.91  & 94.42  & 74.09  & 69.24  & 73.72  \\
    \rowcolor{gray!40}
    \multicolumn{1}{l|}{CraFT} & 91.94  & \underline{93.92}  & \underline{36.89}  & \underline{63.28}  & 72.07  & \underline{78.11}  & 83.66  & \underline{70.97}  & 94.48  & \underline{79.78}  & 68.21  & \underline{75.76}  \\
    \rowcolor{gray!40}
    \multicolumn{1}{l|}{CraFT + vp} & \textbf{93.12}  & \textbf{95.40}  & \textbf{41.17}  & \textbf{66.86}  & \textbf{82.40}  & \textbf{80.95}  & 83.75  & \textbf{72.60}  & \textbf{95.29}  & \textbf{80.83}  & \underline{70.05}  & \textbf{78.40}  \\
    \bottomrule
    \end{tabular}%
    }
  \label{tab:more_experiemnt}%
\end{table}%

\setlength{\tabcolsep}{6pt}
\begin{table}[htbp]
  \centering
  \caption{
    \textbf{The time cost of training these methods.}
    We report the total training time of BBT~\cite{sun2022black}, BlackVIP~\cite{oh2023blackvip}, \citet{yu2023black}, and our methods on the 11 16-shot datasets with 3 random seeds.
    The training time is computed on a single RTX 3090.
  }
    \begin{tabular}{cc|ccccc}
    \toprule
    \multicolumn{2}{c|}{} & \multicolumn{1}{l}{BBT~\cite{sun2022black}} & \multicolumn{1}{l}{BlackVIP~\cite{oh2023blackvip}} & \multicolumn{1}{l}{\citet{yu2023black}} & \multicolumn{1}{l}{CraFT} & \multicolumn{1}{l}{CraFT+vp} \\
    \midrule
    \multicolumn{2}{c|}{Training Time} &   $\sim$ 2 days    &   $\sim$ 1 month    &   $\sim$ 2 weeks   &   $\sim$ 2 days  &  $\sim$ 2 weeks \\
    \bottomrule
    \end{tabular}%
  \label{tab:time_cost}%
\end{table}%

\section{Dataset Statistics}
In the experiments, we selected a set of 15 public datasets to assess the efficacy of our method and the baseline methods.
A comprehensive overview of the datasets, including detailed statistics, is presented in Table~\ref{tab:datasets}.

\setlength{\tabcolsep}{6pt}
\begin{table}[htbp]
  \centering
  \caption{Detailed statistics of datasets used in experiments.}
  \resizebox{0.7\textwidth}{!}{
    \begin{tabular}{lrrrcc}
    \toprule
    Dataset & \# Classes & \# Training & \# Test & \multicolumn{2}{c}{Task} \\
    \midrule
    {OxfordPets} & 37    & 2,944 & 3,669 & \multicolumn{2}{c}{fine-grained pets recognition} \\
    {Flowers102} & 102   & 4,093 & 2,463 & \multicolumn{2}{c}{fine-grained flowers recognition} \\
    {FGVCAircraft} & 100   & 3,334 & 3,333 & \multicolumn{2}{c}{fine-grained aircraft recognition} \\
    {DTD} & 47    & 2,820 & 1,692 & \multicolumn{2}{c}{Textural recognition} \\
    {EuroSAT} & 10    & 13,500 & 8,100 & \multicolumn{2}{c}{Satellite image recognition} \\
    {StanfordCars} & 196   & 6,509 & 8,041 & \multicolumn{2}{c}{Fine-grained car recognition} \\
    {Food101} & 101   & 50,500 & 30,300 & \multicolumn{2}{c}{Fine-grained food recognition} \\
    {Sun397} & 397   & 15,880 & 19,850 & \multicolumn{2}{c}{Scene recognition} \\
    {Caltech101} & 100   & 4,128 & 2,465 & \multicolumn{2}{c}{Object recognition} \\
    {UCF101} & 101   & 7,639 & 3,783 & \multicolumn{2}{c}{Action recognition} \\
    {ImageNet} & 1,000 & 1.28M & 50,000 & \multicolumn{2}{c}{Object recognition} \\
    \midrule
    {ImageNetV2} & 1,000 & -     & 10,000 & \multicolumn{2}{c}{Robustness of collocation} \\
    {ImageNet-Sketch} & 1,000 & -     & 50,889 & \multicolumn{2}{c}{Robustness of sketch domain} \\
    {ImageNet-A} & 200   & -     & 7,500 & \multicolumn{2}{c}{Robustness of adversarial} \\
    {ImageNet-R} & 200   & -     & 30,000 & \multicolumn{2}{c}{Robustness of rendition styles} \\
    \bottomrule
    \end{tabular}%
    }
  \label{tab:datasets}%
\end{table}%

\end{document}